\documentclass{bmvc2k}

\usepackage[utf8x]{inputenc}
\usepackage{graphicx}
\usepackage{wrapfig}
\usepackage{array}
\usepackage{booktabs}
\usepackage{enumitem}   
\usepackage{amssymb}
\usepackage{tabularx}
\usepackage{adjustbox}
\begin{document}

\title{Progressive Face Super-Resolution via Attention to Facial Landmark}

\addauthor{Deokyun Kim*}{deokyunkim@kaist.ac.kr}{1} 

\addauthor{Minseon Kim*}{minseonkim@kaist.ac.kr}{1} 

\addauthor{Gihyun Kwon*}{cyclomon@kaist.ac.kr}{1} 
 
\addauthor{Dae-Shik Kim}{daeshik@kaist.ac.kr}{1}
 

 \addinstitution{
  School of Electrical Engineering,\\
  Korea Advanced Institute
of Science and Technology,\\
Republic of Korea
 }

\runninghead{D. Kim, M. Kim, G. Kwon et al.}{Progressive Face Super-Resolution}

\def\eg{\emph{e.g}\bmvaOneDot}
\def\Eg{\emph{E.g}\bmvaOneDot}
\def\etal{\emph{et al}\bmvaOneDot}



\maketitle
\begin{abstract}

Face Super-Resolution (SR) is a subfield of the SR domain that specifically targets the reconstruction of face images. The main challenge of face SR is to restore essential facial features without distortion. We propose a novel face SR method that generates photo-realistic 8$\times$ super-resolved face images with fully retained facial details. To that end, we adopt a progressive training method, which allows stable training by splitting the network into successive steps, each producing output with a progressively higher resolution. We also propose a novel facial attention loss and apply it at each step to focus on restoring facial attributes in greater details by multiplying the pixel difference and heatmap values. Lastly, we propose a compressed version of the state-of-the-art face alignment network (FAN) for landmark heatmap extraction. With the proposed FAN, we can extract the heatmaps suitable for face SR and also reduce the overall training time. Experimental results verify that our method outperforms state-of-the-art methods in both qualitative and quantitative measurements, especially in perceptual quality.



\end{abstract}

\section{Introduction}


Face Super-Resolution (SR) is a domain-specific SR which aims to reconstruct High Resolution (HR) face images from Low Resolution (LR) face images while restoring facial details. When enlarging the LR face images to high-resolution images, the HR images suffer from face distortion. The finer details of faces disappear incurring misperception of facial attributes on faces. In an attempt to address this problem, the previous studies \cite{Yu2018, ChengHan2018} embedded additional facial attribute vectors into network feature maps to reflect facial attributes in super-resolved face images. These approaches require prior information for face SR; however, the additional information is difficult to obtain in the wild. Other studies incorporate facial landmark information by employing auxiliary networks such as face alignment network \cite{XinYu2018,Adrian2018}, and prior estimation network \cite{Yuchen18}. However, these approaches tend to concentrate on the localization of facial landmarks without sufficient consideration of the facial attributes in the areas around landmarks.

Different from the previous works, we propose a face SR method which restores original facial details more precisely by giving strong constraints to the landmark areas. To stably generate photo-realistic 8$\times$ upscaled images, we adopt a progressive training method \cite{TeroKarras2018, Yifan2018, Wei2017, Namhyuk2018} which grows both generator and discriminator progressively. We also introduce a novel facial attention loss which makes our SR network to restore the accurate facial details. The attention loss is applied in both the intermediate and the last step of our progressive training. 

Constraining the outputs by applying the attention loss at each step, the output images of each step reflect more accurate facial details.
To obtain the attention loss, we extract the heatmaps from the pre-trained Face Alignment Network (FAN). The extracted heatmaps are used as weights of the pixel difference of the adjacent areas to the landmarks.
Instead of using the state-of-the-art FAN \cite{Adrian2017}, we suggest a compressed network of FAN, called distilled FAN, which is trained by a hint-based method \cite{Adriana15}. The distilled FAN delivers comparable performance to the original FAN while being much more compact. With our approach, we obtain SR-oriented landmark heatmaps as well as significantly reduce the overall training time. Therefore, our method generates super-resolved face images which successfully reflect accurate details of facial components.

For the evaluation, we measure the performance of our method on both aligned and unaligned face images from CelebA \cite{ziwei2015} and AFLW \cite{koestinger2011annotated} datasets. To compare the quality of our results, we calculate the conventional measurements of the average Peak Signal to Noise Ratio (PSNR), Structural SIMilarity (SSIM) \cite{wang2004}, and Multi-Scale Structural SIMilarity (MS-SSIM) \cite{Wang2003}. By conducting an ablation study, we verify that the proposed loss function is beneficial to super-resolving LR face images; we demonstrate the superiority of our method by comparing the results with those of previous studies. We further conduct Mean-Opinion-Score (MOS) \cite{christrian16} test to measure the perceptual quality. The experimental results show that our network successfully generates high-fidelity face images, accurately preserving the original features around the facial landmarks. 
In summary, our contributions are as follows:
\begin{enumerate}[nosep]



\item To the best of our knowledge, progressive training method is used in natural image SR, but this is the first method which leverages the progressive training method for face SR. We give constraints to each step of the SR network and generate high-quality face images reflecting details of facial components. 

\item Facial attention loss makes the SR network learn to restore facial details with the method of focusing on the adjacent area of facial landmarks, which is verified by our super-resolved results. 



\item We compress the state-of-the-art FAN into a smaller network using hint-based method. With the distilled FAN, we are able to extract meaningful landmark heatmaps which are more suitable for a face SR task and reduce the overall training time.



\end{enumerate}





\section{Related work}

Utilizing facial information, such as facial attributes and spatial configuration of facial components, is the key factor in face SR. Yu et al. \cite{Yu2018} interweave multiple spatial transformer networks to satisfy the requirement of face alignment as well as embeds facial attribute vectors to lower the ambiguity in facial attributes. Lee et al. \cite{ChengHan2018} fuses the information of both image domain and attribute domain in order to reflect facial attributes in super-resolved images. These methods preserve facial attributes indicated by facial attribute vectors. However, attribute vectors are not only difficult to acquire in the wild but also limited to describing partial facial attributes.


While preserving the facial landmarks location, Chen et al. \cite{Yuchen18} propose the face shape prior estimation network, which provides a solution for accurate geometry estimation obtained from coarse HR face images. Yu et al. \cite{XinYu2018} estimate the spatial position of their facial components to preserve the original spatial structure in upscaled face images. In addition, Bulat et al. \cite{Adrian2018} propose a heatmap loss to localize landmarks in super-resolved images so as to upscale input face images $4\times$. Although these methods preserve the spatial configuration of facial components, they fail to fully reflect accurate facial attributes.
In contrast to the previous works, our method carefully considers the facial attributes around the landmarks to restore the facial details without prior information as well as preserves landmark location.

\section{Approach}
\begin{figure}[t!]
\centering
\includegraphics[width=1.0\textwidth]{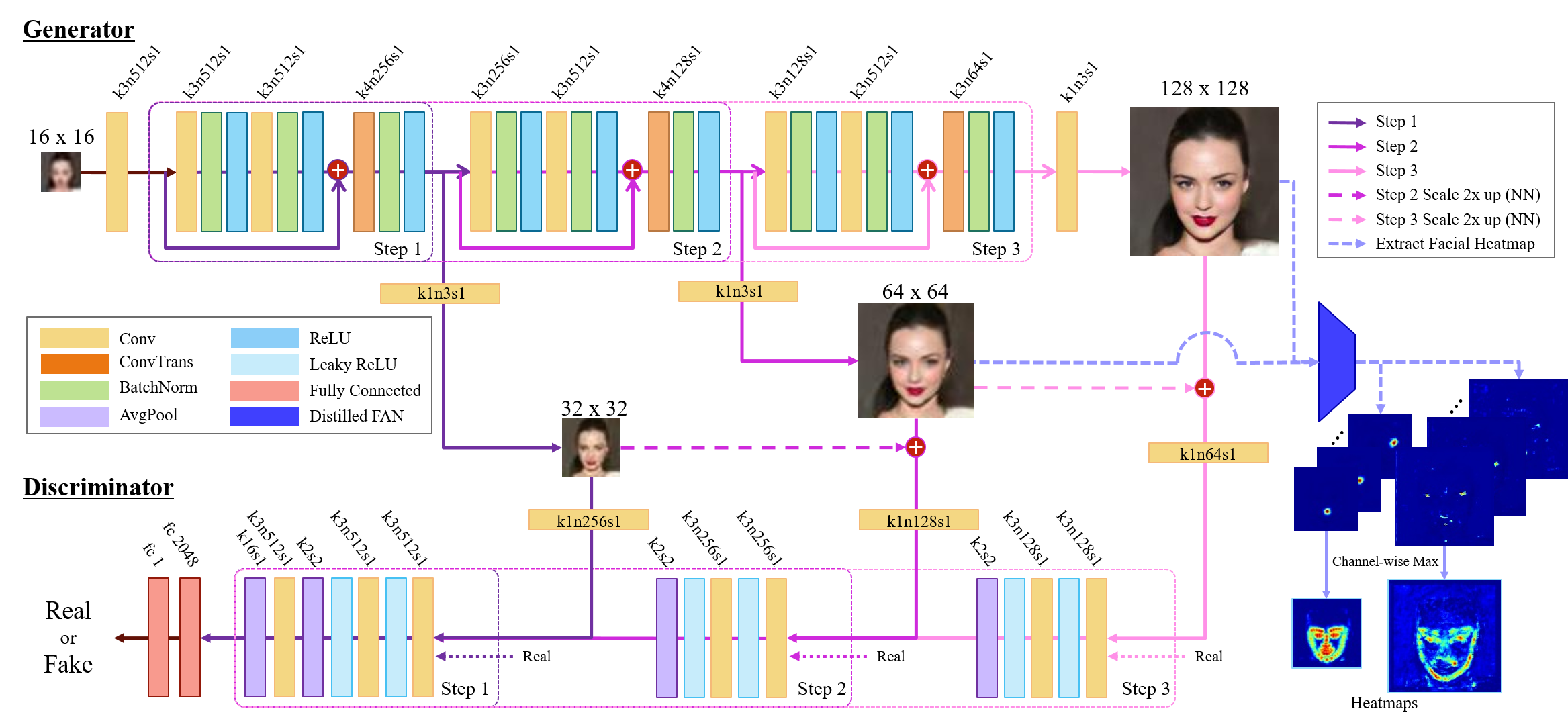}
\caption{Our network architecture overview. \textit{(k : kernel size, n : output channel, s : stride)}}
\label{fig:architecture}
\end{figure}

In this section, we describe our methods for the enhanced face SR. To generate the high-fidelity super-resolved face images that reflect the facial attributes of target face images, three main approaches are used: progressive training, facial attention loss, and distillation of Face Alignment Network (FAN).

\subsection{Progressive Face SR Network}
The overview of our network architecture is shown in Figure \ref{fig:architecture}. To incorporate the adversarial loss, our architecture is composed of the generator network, which is our face SR network, and the discriminator network. To train the generator and the discriminator stably, we construct both the networks which consist of layers stacked by steps. 
Our generator network consists of three residual blocks \cite{Kaiming15} with batch normalization layers (BatchNorm), transpose convolution layers (ConvTrans), and Rectifier Linear Unit (ReLU) as an activation function. The discriminator network has a corresponding architecture to the generator network, which is comprised of convolution layers (Conv), average pooling layers (AvgPool), and Leaky ReLU. To improve the discriminator performance, we calculate the standard deviation of input batch, then replicate the value into a one-dimensional feature map, and concatenate it to the end of the discriminator \cite{TeroKarras2018}. We use additional convolution layers in each step in order to convert the intermediate feature maps into RGB images, and vice versa.

In Step 1, each network employs one block for training and learns to upscale images $2\times$. These 2$\times$ upscaled outputs from the generator go through the corresponding part of the discriminator, and the outputs are then compared with target images. In Step 2, the $2\times$ upscaled outputs from Step 1 are upscaled $2\times$ again by nearest-neighbor interpolation, and then the interpolated outputs are added to $4\times$ upscaled outputs from Step 2. This process is expressed as follows: $(1-\alpha)*f(G^{N-1}(I)) + \alpha *G^{N}(I)$, where $G$ is our SR network, $f$ is nearest neighbor (NN) interpolation, $I$ and $N\in \left\{ 2,3 \right\}$ denote input images and number of step, respectively. A weight scale $\alpha$ increases linearly from zero to one. The upscaled outputs are compared to the corresponding target images through the discriminator. The same procedure above is implemented in Step 3 (8$\times$). The method allow the network learn super-resolving face images with different loss in each step effectively and stably.

\subsection{Facial Attention Loss} 
We propose the \textit{facial attention loss} to restore the attributes of the adjacent area to the facial landmarks. This \textit{facial attention loss} makes the face SR network focus on the facial details around the predicted landmark area by element-wise multiplying landmark attention heatmaps $M^*$, and the $L1$ distance between the upscaled images and the corresponding target images. To achieve this, we employ facial landmark heatmaps which contain landmark location information. The \textit{facial attention loss} is defined as:

\begin{equation}
    L_{attention} = \frac{1}{r^2WH}\sum\limits_{x=1}^{rW}\sum\limits_{y=1}^{rH}(M^*_{x,y}\cdot|I^{HR}_{x,y}-G(I^{LR} )_{x,y}|)
\end{equation}
where $G$ is face SR network, $I^{HR}$ and $I^{LR}$ are target face images and input LR images, respectively. The landmark attention heatmap $M^*$ is channel-wise max values of the target heatmap $M$ generated from target face images. To compensate for the variance between the landmarks, the heatmap $M$ is min-max normalized into [0,1]. The heatmap $M$ has the dimension of $N \times rW \times rH$, where $N$ is number of landmarks, $W$ and $H$ are width and height of the input image. The upscale factor $r$ is set to be 4 and 8 in Step 2 and Step 3, respectively. To give attention at images with enough information, we adjust facial attention loss at upscaled outputs with size of 64$\times$64 and 128$\times$128.

\subsection{Distilled Face Alignment Network}

\begin{wrapfigure}{L}{0.4\textwidth}
\centering
\includegraphics[width=0.4\textwidth]{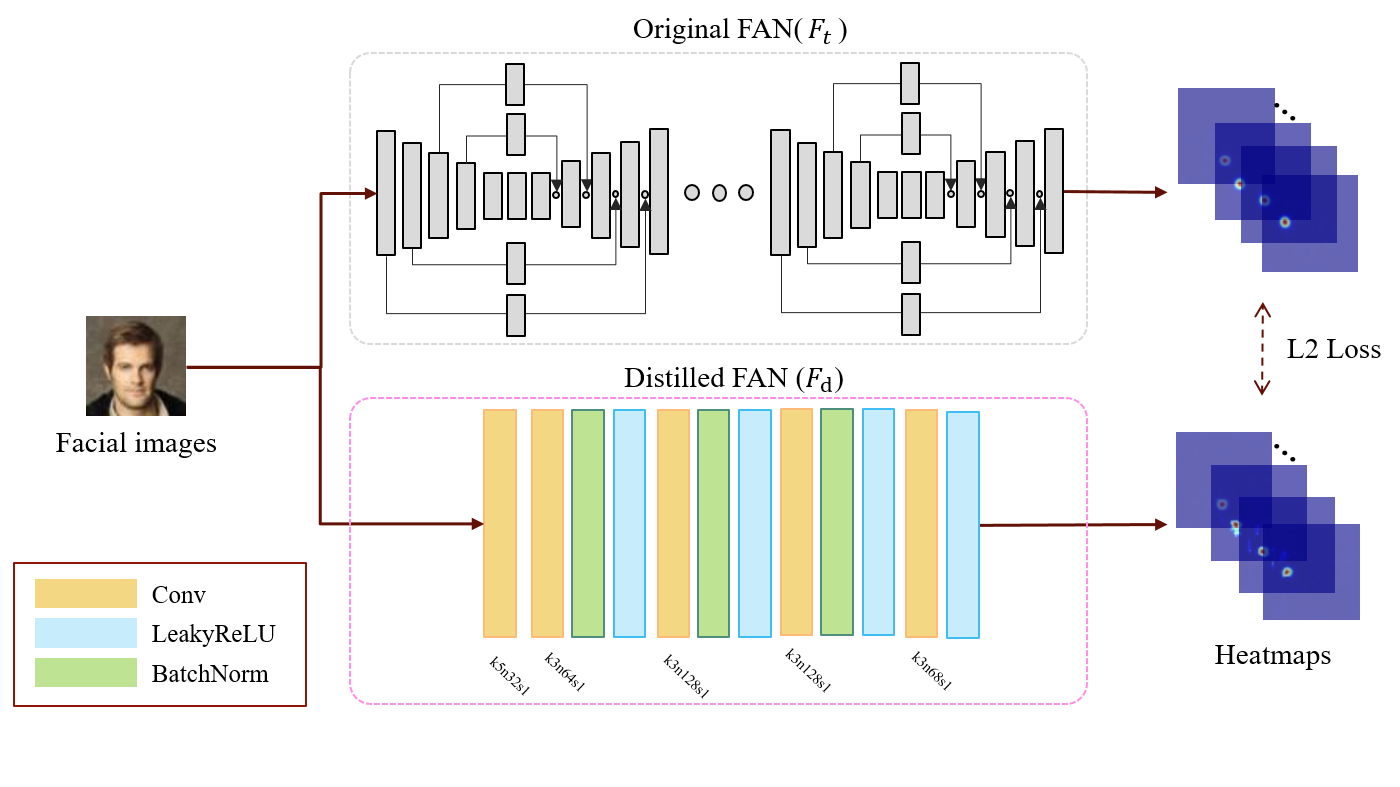}
\caption{Distilled face alignment network hint based training}
\label{fig:distilledFAN}
\end{wrapfigure}
The state-of-the-art FAN \cite{Adrian2017} predicts the location of all landmarks including occluded landmarks exploiting multiple-scale feature maps from four-stacked Hourglass architecture which consists of encoder-decoder and skip-layer\cite{Alejandro2016}. This approach predicts landmark locations based on heatmap values, which are highly concentrated around the landmark points. However, for face SR, we aim to give attention to the overall facial landmark area in the images except for the occluded landmark area.

To generate heatmaps suitable for giving attention to accurate facial landmark area, we construct the network with neither encoder-decoder architecture nor skip-layer so as to predict landmarks based on single-scale feature maps. Also, in order to reduce overall training time and achieve comparable performance to state-of-the-art FAN, we compress the FAN into the network shown in Figure \ref{fig:distilledFAN} based on a hint-based training method \cite{Adriana15}. We train the distilled FAN to minimize $\sum \left( F_{d}\left( I \right) - F_{t}\left( I \right) \right)^{2}$, where $F_d$ and $F_t$ represents our distilled FAN and original FAN, $I$ denotes input face images. This approach has two advantages: it provides face SR-oriented heatmaps, and it reduces the overall training time from $\sim 3$ days to $\sim 1$ days in our experiments.

\subsection{Overall Training Loss} 
\noindent\textbf{MSE loss} 
We use the pixel-wise Mean-Square-Error (MSE) loss to minimize the distance between the HR target image and the super-resolved image.

\begin{equation}
    L_{pixel} = \frac{1}{r^2WH} \sum_{x=1}^{rW}\sum_{y=1}^{rH}\left(I_{x,y}^{HR} -G\left(I^{LR}  \right)_{x,y} \right)^{2}
\end{equation}
\noindent\textbf{Perceptual loss} 
A perceptual loss \cite{Justin16} is proposed to prevent generating blurry and unrealistic face images, and to obtain more realistic HR images. The loss over the pre-trained VGG16  \cite{Karen2015} features at a given layer $i$ is defined as:

\begin{equation}
    L_{feat/i} = \frac{1}{W_{i}H_{i}}\sum_{x=1}^{W_i}\sum_{y=1}^{H_i} \left(\phi_i \left(I^{HR} \right)_{x,y} - \phi_i\left(G\left(I^{LR} \right) \right)_{x,y}  \right)^2
\end{equation}
where $\phi_i$ denotes the feature map obtained after the last convolutional layer of the i−th block.

\noindent\textbf{Adversarial loss} We use the WGAN Loss \cite{Martin2017} to stabilize the training process. In WGAN, the loss function is defined as the Wasserstein distance between the distribution of target $I^{HR} \sim P_r$ and those of the generated images $\Tilde{I} \sim P_g$. For further improvement of training stability, we apply the Gradient Penalty term proposed in WGAN-GP \cite{XiangWei2018}, which enforces the Lipschitz -1 condition of the discriminator.
$\hat{I}$ is a randomly sampled image among the samples from $P_r$ and $P_g$.
Therefore, the loss function is as follows:

\begin{equation}
    L_{WGAN} = 
    \mathbb{E}_{I^{HR} \sim P_r}{[D(I^{HR})]} - \mathbb{E}_{ \Tilde{I} \sim P_g}{[D( \Tilde{I})]} + \lambda \mathbb{E}_{\hat{I} \sim P_{\hat{I}}}[|| \nabla_{\hat{I}}D(\hat{I})_2 -1||^2]
\end{equation}

\noindent\textbf{Heatmap Loss} As proposed by \cite{Adrian2018}, the heatmap loss improves the structural consistency of face images by minimizing the distance between the heatmaps of both generated images and target ones. The heatmap loss function is described as:

\begin{equation}
    L_{heatmap} = \frac{1}{r^2NWH}\sum\limits_{n=1}^{N}\sum\limits_{x=1}^{rW}\sum\limits_{y=1}^{rH}(M^n_{x,y}-\Tilde{M}^n_{x,y})^2
\end{equation}
where $N$ is the number of landmarks, $M$ and $\Tilde{M}$ are calculated as $M=F_{d}(I^{HR})$ and $\Tilde{M}=F_{d}(G\left(I^{LR}  \right))$.

\noindent\textbf{Overall training loss} Since the landmark losses are applied to Step 2 \& 3, we intend not to include the $L_{heatmap}$ and $L_{attention}$ in Step 1. The final loss term is shown as:

\begin{equation}
      \begin{gathered}
         L_{Ours} = \alpha L_{pixel} + \beta L_{feat} + \gamma L_{WGAN}\mbox{,{\textit{   at step 1}}} \\
        L_{Ours} = \alpha L_{pixel} + \beta L_{feat} + \gamma L_{WGAN} + \lambda  L_{heatmap} + \eta L_{attention}\mbox{,{\textit{   at step 2 \& step3}}} \\
    \end{gathered}
\end{equation}
where $\alpha$, $\beta$, $\gamma$, $\lambda$ and $\eta$ are the corresponding weights.

\section{Experiments}
\subsection{Implementation details}
\noindent\textbf{Datasets}
In our experiments, we use two different datasets: aligned dataset and unaligned one. The aligned CelebA dataset \cite{ziwei2015} is used to test how accurately the facial details can be restored. The unaligned CelebA and AFLW \cite{koestinger2011annotated} datasets are used to verify the applicability of our face SR network in real world. The aligned face images are cropped into square. The face images of the unaligned dataset are cropped based on the bounding box areas. The cropped images are resized into 128$\times$128 pixels to be used as targets of Step 3, and bilinearly downsampled into 64$\times$64 pixels as targets of Step 2, 32$\times$32 pixels as targets of Step 1, and 16$\times$16 pixels as LR inputs. We use all 162,770 images as a training set, and 19,867 images as a test set from aligned CelebA dataset. For the unaligned dataset, we use 80,000 cropped images from unaligned CelebA, and 20,000 from AFLW for training. As a test set, 5,000 images from CelebA and 4,384 images from AFLW are used.


\noindent\textbf{Training details}
We implement our face SR network using PyTorch \cite{paszke2017}. We train our networks using the Adam optimizer \cite{Diederik15} with a learning rate of $1\times10^{-3}$, and the mini-batch size of 16. The training iteration of each step is set by hyperparameter. In our model, we train our model 50K, 50K and 100K iterations, empirically. Running totally 200K iterations takes $\sim$1 day on single Titan X GPU. In addition, we train the distilled FAN using the Adam optimizer with a learning rate of $1\times10^{-4}$, mini-batch size of 16, and 100K iterations.

\begin{figure}[t]
\centering
\includegraphics[width=\textwidth]{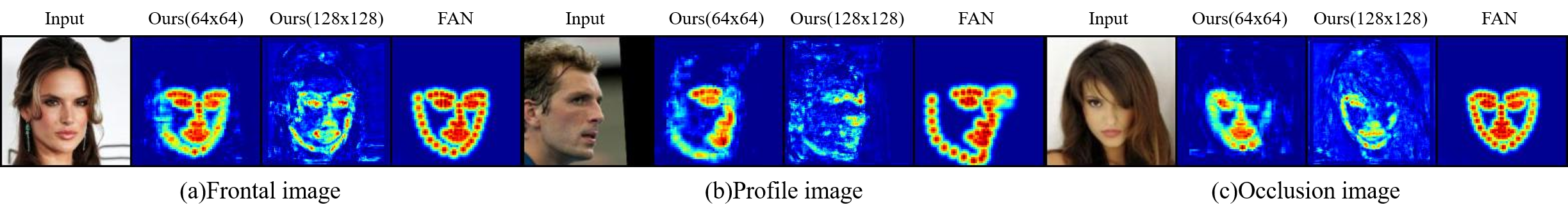}
\caption{Distilled FAN results(Ours) comparison with FAN results \cite{Adrian2017}.}
\vspace*{-4mm}
\label{fig:heatmap}
\end{figure}

\begin{figure}[t]
\centering
\includegraphics[width=\textwidth]{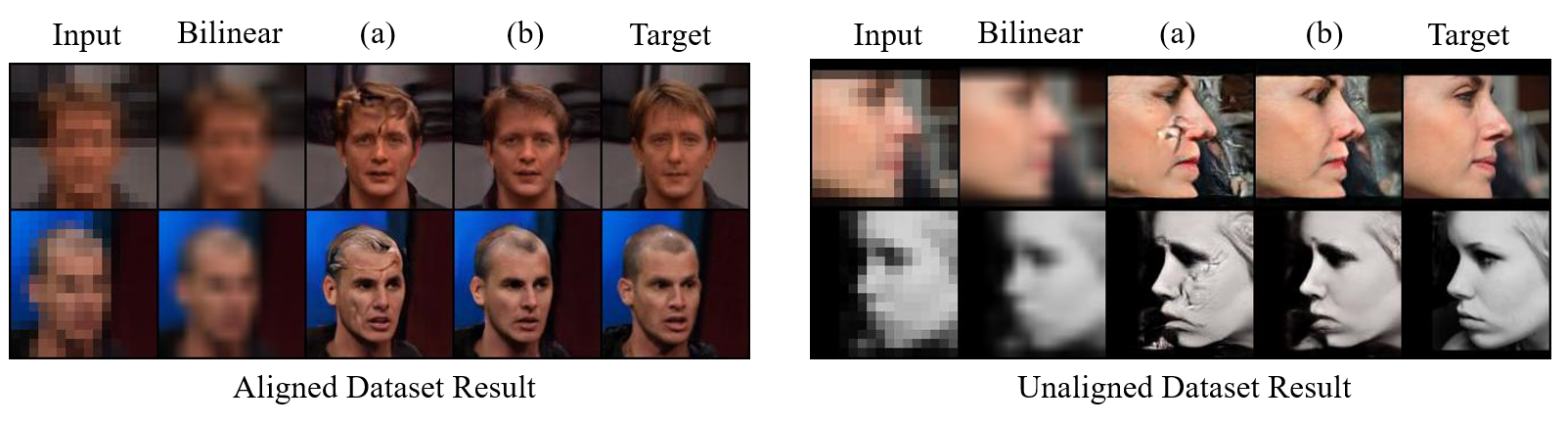}
\caption{ Our image results (a) with the original FAN, (b) with the distilled FAN.}
\vspace*{-4mm}
\label{fig:distilledresults}
\end{figure}

\subsection{Distilled FAN Results}





In this section, we compare our distilled FAN to the original FAN \cite{Adrian2017}. To verify how similarly the distilled FAN predicts landmark location compared to the original FAN, we use Normalized Mean Error (NME) metric \cite{Adrian2017, Xiangxin2012} between the predicted landmark locations from the distilled and the original FAN. The NME is calculated as $ NME = \frac{1}{N}\sum_{k=1}^{N}\frac{\parallel g_k - p_k \parallel_2 }{d}$, where $g$ denotes the landmark from original FAN, $p$ is the corresponding prediction from the distilled model, and $d$ is the facial image size. The NME evaluation results and the number of parameters are shown in Table \ref{table:distlledcomparison}. The results show that our distilled FAN predicts facial landmarks with comparable performance, and it has much fewer parameters than the original FAN. 

As shown in Figure \ref{fig:heatmap}(b) and (c), the output heatmaps of the original FAN have high values around the landmark points even in the occluded area, but the output heatmaps of distilled FAN have relatively low values in the occluded landmark area. The heatmaps of our distilled FAN are suitable for facial attention weights.



Figure \ref{fig:distilledresults}(a) shows the problem of using the heatmap from the original FAN as attention weights. There is no significant distortion in facial attributes, but it leads to some artifacts because the attention is applied only to the distinct points of facial landmarks. As shown in Table \ref{table:distlledcomparison}, the distilled FAN improves SR performance with a small number of parameters.






\begin{table*}[ht]
\begin{center}
\begin{adjustbox}{width=\textwidth}
\begin{tabular}{@{\extracolsep{\fill}}c||c||cc|ccc|cc|ccc}
{}& Parameters&\multicolumn{5}{c|}{Aligned} & \multicolumn{5}{c}{Unaligned}\\
 &(ratio)& \multicolumn{2}{c|}{NME} &PSNR&SSIM&MS-SSIM&\multicolumn{2}{c|}{NME} &PSNR&SSIM&MS-SSIM\\
\hline \hline
\begin{tabular}{@{}c@{}}FAN \cite{Adrian2017} \\ (original)\end{tabular} &\begin{tabular}{@{}c@{}}23,820,176 \\ (100\%)\end{tabular}&\begin{tabular}{@{}c@{}}(64x64) \\ -\end{tabular} &\begin{tabular}{@{}c@{}}(128x128) \\ -\end{tabular}  &22.29 & 0.670 & 0.895 &\begin{tabular}{@{}c@{}}(64x64) \\ -\end{tabular} &\begin{tabular}{@{}c@{}}(128x128) \\ -\end{tabular} &22.51 & 0.667& 0.886\\
\hline 

\textbf{Ours} &\begin{tabular}{@{}c@{}}321,412 \\ (1.35\%)\end{tabular}& 0.239\% & 0.987\%&\textbf{22.66}&\textbf{0.685} &\textbf{0.902}& 0.830\% &2.2643\%  &\textbf{22.96} &\textbf{0.695}&\textbf{0.897}\\

\end{tabular}
\end{adjustbox}
\end{center}
\caption{Parameters, NME evaluation, PSNR, SSIM, and MS-SSIM comparison results}
\label{table:distlledcomparison}
\end{table*}




\subsection{Ablation Study}
In order to observe the effects of each element of our method, we conduct an ablation study using the conventional measurements of the average PSNR, SSIM, and MS-SSIM; Minimizing the MSE maximize the PSNR, which is commonly used to evaluate the SR results. Since PSNR is defined based only on pixel-wise differences, the value of PSNR has limitation to represent perceptually relevant differences \cite{christrian16}. Therefore, we further measure SSIM and MS-SSIM.

\begin{figure}[t]
\centering
\includegraphics[width=0.68\textwidth]{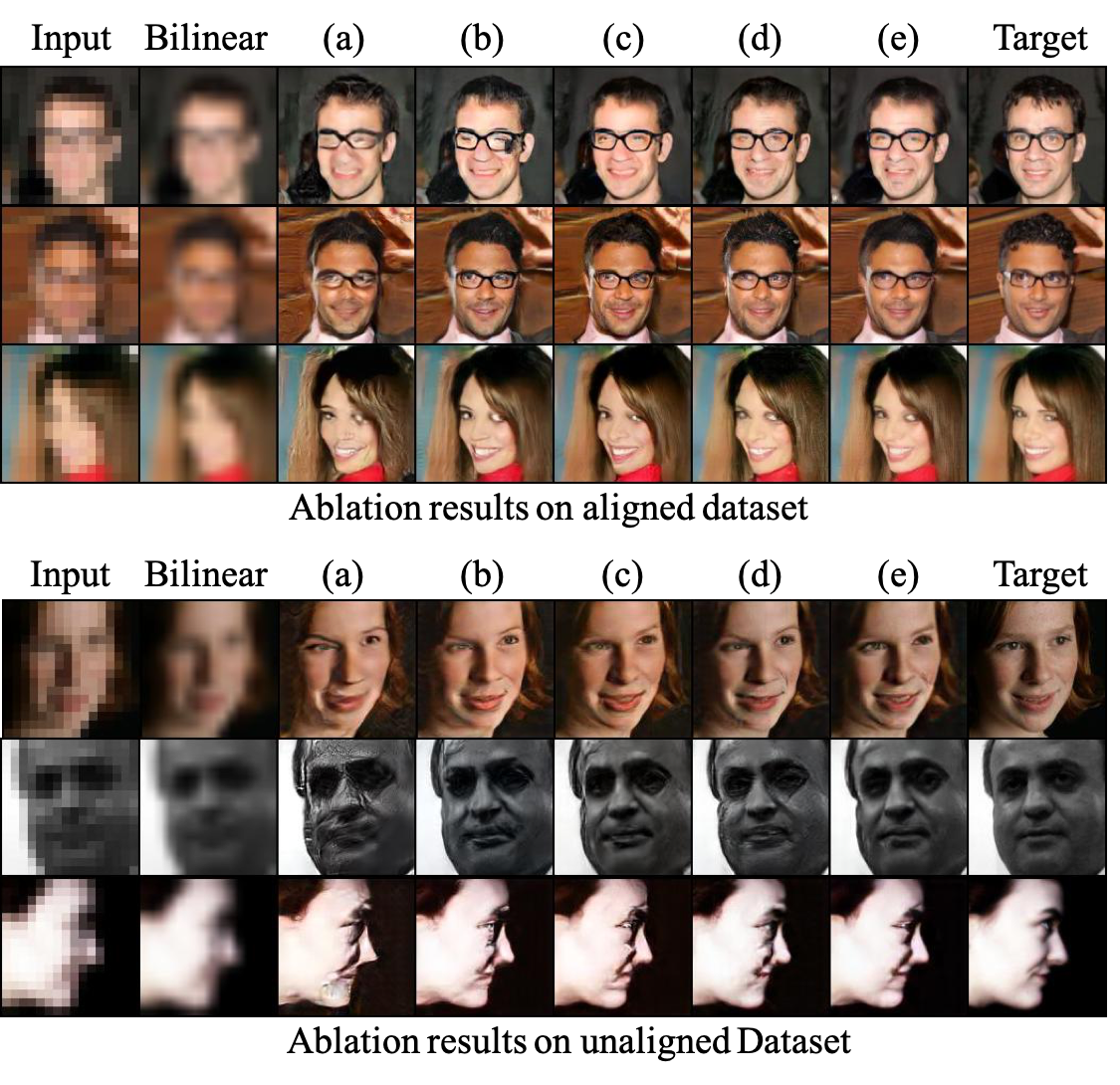}

\caption{Ablation study results on aligned and unaligned datasets. (a) $L_{pixel}+L_{WGAN}$ (b) $L_{pixel}+L_{WGAN}+L_{feat}$ (c) $L_{pixel}+L_{WGAN}+L_{feat}+L_{hm}$ (d) $L_{Ours}$-\textit{no progressive} (e) $L_{Ours}$.}
\vspace*{-2mm}
\label{fig:ablation}
\end{figure}

\begin{table*}[t]
\begin{center}
\begin{adjustbox}{width=0.9\textwidth}
\begin{tabular*}{\textwidth}{@{\extracolsep{\fill}}c||ccc|ccc}
{}& \multicolumn{3}{c|}{Aligned} & \multicolumn{3}{c}{Unaligned}\\
Method& PSNR &SSIM& MS-SSIM&PSNR&SSIM&MS-SSIM \\
\hline \hline

$L_{pixel}+L_{WGAN}$& 21.62 & 0.616 & 0.873 &21.62&0.626&0.863\\

$L_{pixel}+L_{WGAN}+L_{feat}$& 21.89 & 0.649& 0.887 &22.26&0.663&0.884\\

$L_{pixel}+L_{WGAN}+L_{feat}+L_{hm}$& 21.95 &0.650& 0.890 &22.53&0.679&0.892\\
\hline
$L_{Ours}$-\textit{no progressive} &22.24&0.660&0.893&22.83&0.680&0.895\\
\textbf{$L_{Ours}$} & \textbf{22.66} & \textbf{0.685} & \textbf{0.902}&\textbf{22.96}&\textbf{0.695}&\textbf{0.897}\\

\end{tabular*}
\end{adjustbox}
\end{center}
\caption{PSNR, SSIM and MS-SSIM values for the ablation study results on aligned and unaligned datasets.}
\label{table:ablation}

\end{table*}

\noindent\textbf{Effects of loss functions}
We conduct three experiments to estimate the effect of perceptual loss, heatmap loss, and proposed \textit{facial attention loss} on the aligned dataset and the unaligned dataset. 
Figure \ref{fig:ablation} shows the results of using different loss functions on both aligned and unaligned datasets. The result images without perceptual loss have severely deteriorated texture of face images. Moreover, the result images without heatmap loss have unclear shapes and distortion around the eyes and mouths. As the \textit{facial attention loss} uses the landmark heatmaps as guidance, it is helpful for face images to restore facial details.

As shown in Table \ref{table:ablation}, using our \textit{facial attention loss} shows the highest value in PSNR, SSIM and MS-SSIM. These results verify that our \textit{facial attention loss} is helpful to generate more structurally meaningful face images.

\noindent\textbf{Effects of progressive training} 
To verify that the progressive training method is helpful to super-resolve face images, we train our face SR network without the progressive training method. The qualitative comparison of outputs is shown in Figure \ref{fig:ablation}(d).
There is some degradation of facial details in super-resolved images from our non-progressively trained network, which is trained by minimizing $L_{ours}$. As shown in Table \ref{table:ablation}, the measurement values are also increased by progressive training method.



\begin{figure}[t]
\centering
\includegraphics[width=0.7\textwidth]{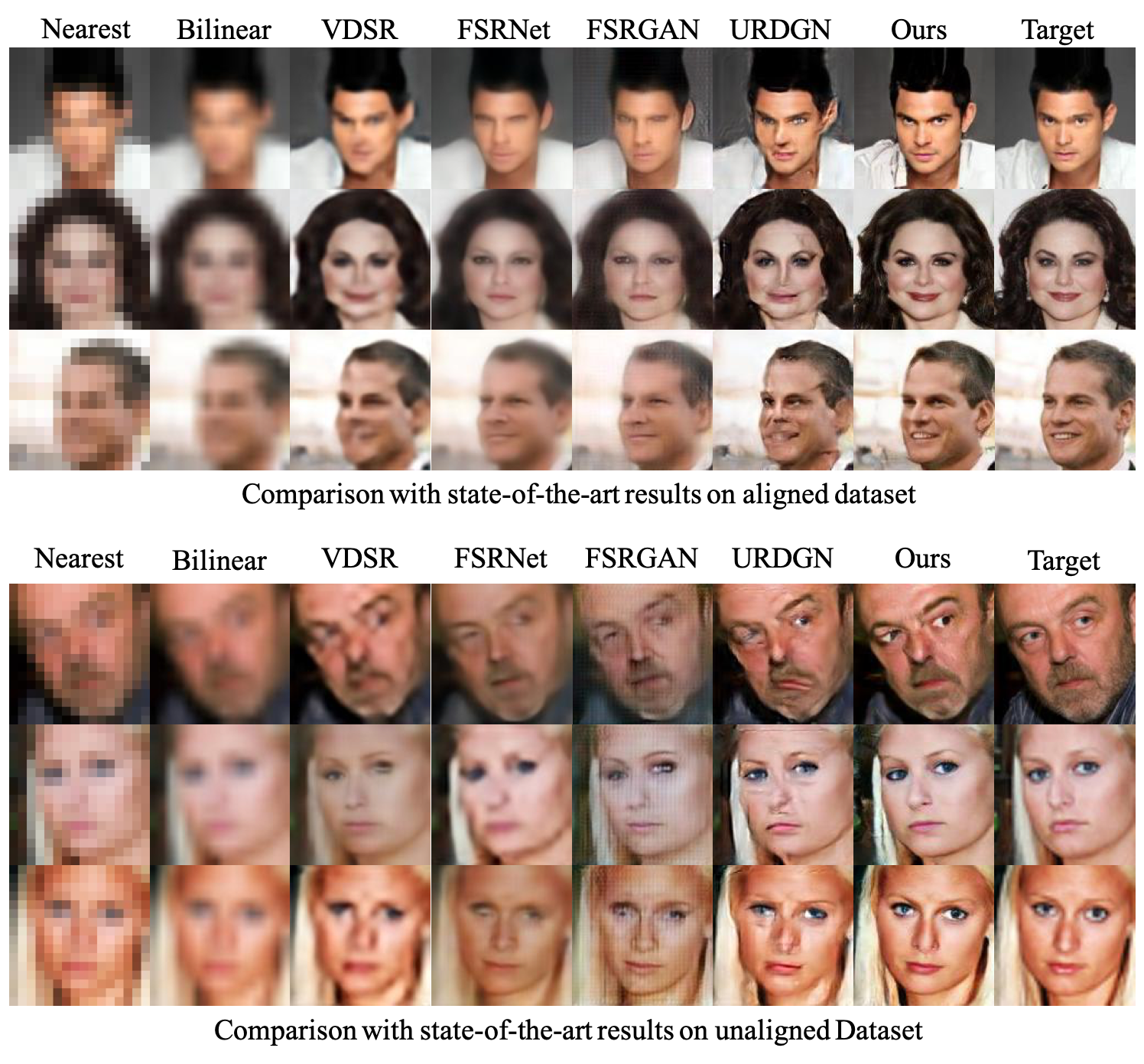}

\caption{Qualitative comparison with aligned and unaligned datasets}
\vspace*{-2mm}
\label{fig:sota}
\end{figure}

\begin{table*}[t]
\begin{center}
\begin{adjustbox}{width=\textwidth}
\begin{tabular}{@{\extracolsep{\fill}} c |c c  c  c | c c c  c } 
\centering &\multicolumn{4}{c|}{Aligned} &\multicolumn{4}{c}{Unaligned} \\
\centering & PSNR & SSIM & MS-SSIM & MOS & PSNR & SSIM & MS-SSIM & MOS\\
\hline \centering

Bilinear & 20.75 & 0.574 & 0.782 &1.72&21.86 & 0.624 & 0.795&1.52\\
URDGN \cite{XinYu2016}  & 21.99 & 0.622 & 0.875 &2.55 & 23.01 & 0.643 & 0.874 &2.45\\
FSRGAN \cite{Yuchen18}  & 22.27 & 0.601 & 0.841 &2.46 &20.95 & 0.515 & 0.741 &2.28\\
FSRNet \cite{Yuchen18}  & 22.62 & 0.641 & 0.847 &2.34 &21.19 &  0.607 & 0.760&2.19\\
VDSR \cite{Jiwon2016}  & \textbf{22.94} & 0.652 & 0.880 &2.10 & \textbf{23.70}& 0.682 & 0.882 &1.89\\
Ours  & 22.66 & \textbf{0.685} & \textbf{0.902}&\textbf{3.73} & 22.96 &\textbf{0.695}&\textbf{0.897}&\textbf{3.73} \\

\end{tabular}
\end{adjustbox}
\end{center}
\caption{PSNR, SSIM and MS-SSIM values for the baseline experimental results on aligned and unaligned datasets.}
\label{table:sota}

\end{table*}

\subsection{Comparison with State-of-the-Art}
We compare our face SR method to the state-of-the-art SR methods both quantitatively and qualitatively. VDSR \cite{Jiwon2016} employs a pixel-wise $L2$ loss in training. FSRNet and FSRGAN \cite{Yuchen18} employ a facial parsing map in training. URDGN \cite{XinYu2016} employs a spatial convolution layer and a discriminator.

Figure \ref{fig:sota} provides results of various models, and Table \ref{table:sota} presents the quantitative comparisons on the test set. The images from VDSR achieve the highest PSNR, but the results are significantly blurred. As we can see, the results of FSRNet and FSRGAN have realistic features in facial details, but they have artifacts and partially blurred facial components. The URDGN produces relatively clear images but generates distorted face images. 
The results show that our method outperforms other methods especially on SSIM and MS-SSIM, and generates photo-realistic face images with restoring accurate facial attributes. More image results are shown in the Supplementary Materials.

We also conduct a MOS test to quantify image quality based on human vision. We asked 26 raters to assign a score from 1 (bad quality) to 5 (excellent quality) to all the super-resolved images and the high-resolution target images. The raters were calibrated on the 20 images of Nearest Neighbor (score 1) and  HR (score 5) before the main test. The raters rated 8 versions of each image on aligned and unaligned dataset: Nearest neighbor, Bilinear, URDGN, FSRNet, FSRGAN, VDSR, Ours, and HR images (GT). Each rater rated randomly presented 240 images with each dataset (total 480 images). More details are explained in the Supplementary Materials. In Figure \ref{fig:mos}, our method shows overwhelming performance on MOS test, which indicates that our results are perceptually superior to other results.
\section{Conclusion}

\begin{figure}[t!]
\centering
\includegraphics[width=\textwidth]{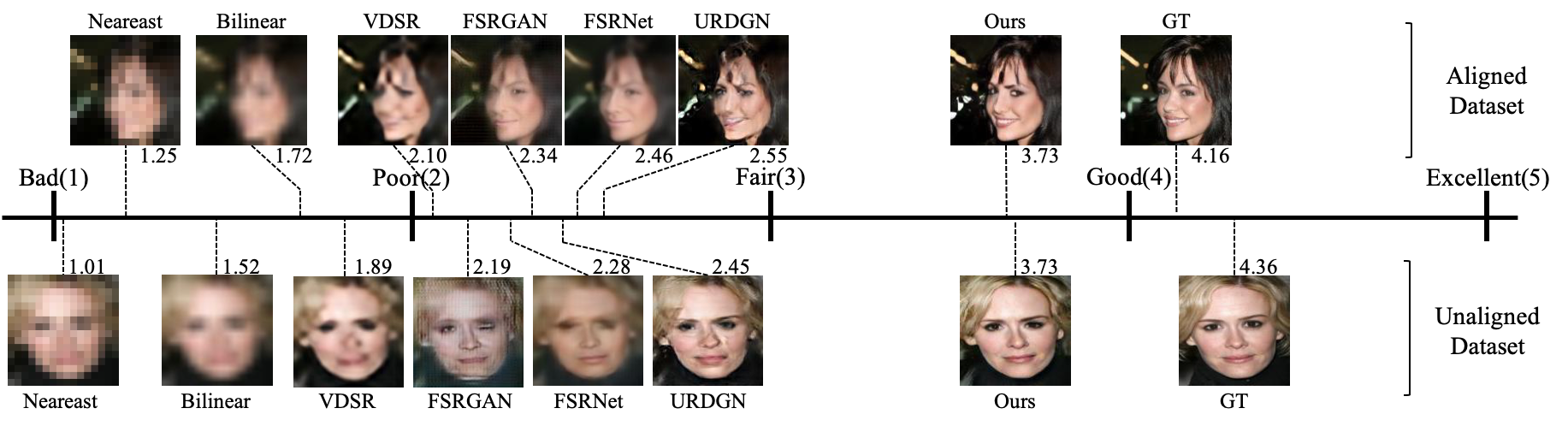}

\caption{MOS result with aligned and unaligned datasets}

\label{fig:mos}
\end{figure}

We propose a novel face SR method which fully reflects facial details. To achieve this, we adopt progressive training method to generate photo-realistic face images and learn restoration of facial details with different guidance in each step. In addition, we propose a new facial attention loss which gives large weights to facial features in the adjacent area of landmarks. Therefore, the facial details are well expressed in super-resolved images. However, the original FAN produces landmark heatmaps including occluded landmark area, which results in degradation of super-resolution performance. Therefore, we suggest distillation of face alignment network to produce more suitable heatmaps for the SR. Besides, our distilled face alignment network has relatively light-weight architecture, so the overall training time is reduced from $\sim 3$ days to $\sim 1$ day. Our experiments demonstrate that our proposed method restore more accurate facial details. In particular, our method produces high-quality face images which are perceptually similar to the real images. As a summary, the proposed method allows our face SR network to super-resolve face images with more precise facial details.

We give attention to specific areas on the faces and propose a method to obtain the heatmaps suitable for face SR. If a better method is developed to obtain the heatmaps which well represent facial landmark areas, we will be able to achieve even better performance through our proposed method. Since our approach restores lost information by focusing on specific areas, we will further be able to restore the desired information by applying our mechanism to any task, such as super-resolution on the medical image, satellite image, and microscopic image, which requires restoration of lost information using super-resolution.

\section{Acknowledgements}

This work was supported by Engineering Research Center of Excellence (ERC) Program supported by National Research Foundation (NRF), Korean Ministry of Science \& ICT (MSIT) (Grant No. NRF-2017R1A5A1014708), Institute for Information \& communications Technology Promotion(IITP) grant funded by the Korea government(MSIT) (No.2016-0-00563, Research on Adaptive Machine Learning Technology Development for Intelligent Autonomous Digital Companion) and Hyundai Motor Company through HYUNDAI-TECHNION-KAIST Consortium.

{\small
\bibliographystyle{ieee}
\bibliography{egbib}
}
\end{document}